%% file: mm.tex
\documentclass[sigconf]{acmart}
\renewcommand\footnotetextcopyrightpermission[1]{}
\usepackage{bbding}
\usepackage{multirow}
\usepackage{makecell}
\usepackage{subfig}
\usepackage{graphicx}
\usepackage{lscape}
\usepackage{enumitem}
\usepackage{colortbl}
\usepackage{booktabs}
\usepackage{adjustbox}
\usepackage{appendix} % 处理附录的包

\usepackage[linesnumbered, ruled]{algorithm2e}
\SetKwRepeat{Do}{do}{while}%
\AtBeginDocument{%
  \providecommand\BibTeX{{%
    \normalfont B\kern-0.5em{\scshape i\kern-0.25em b}\kern-0.8em\TeX}}}

\acmConference[MM'25]{Make sure to enter the correct
  conference title from your rights confirmation email}{October 27 - October 31,
  2025}{Dublin, Ireland.}
\begin{document}


\title{Show and Polish: Reference-Guided Identity Preservation in \\Face Video Restoration}
%Cognitive Style/Personality/Cognitive Pattern

% \author{
% Wenkang Han,
% Wang Lin,
% Yiyun Zhou,
% Qi Liu,
% Shulei Wang,
% Chang Yao,
% Jingyuan Chen\textsuperscript{*}
% \\
% % Zhejiang University \\
% %  wenkangh@zju.edu.cn, jingyuanchen@zju.edu.cn
% \fontsize{12pt}{12pt}\selectfont \textnormal{Zhejiang University}\\
% \fontsize{12pt}{12pt}\selectfont \textnormal{wenkangh@zju.edu.cn}, \fontsize{12pt}{12pt}\selectfont \textnormal{jingyuanchen@zju.edu.cn}
% }

\author{Wenkang Han}
\affiliation{%
  \institution{Zhejiang University}
  % \streetaddress{1 Th{\o}rv{\"a}ld Circle}
  \city{Hangzhou}
  \country{China}
}
\email{wenkangh@zju.edu.cn}

\author{Wang Lin}
\affiliation{%
  \institution{Zhejiang University}
  % \streetaddress{1 Th{\o}rv{\"a}ld Circle}
  \city{Hangzhou}
  \country{China}
}
\email{linwanglw@zju.edu.cn}

\author{Yiyun Zhou}
\affiliation{%
  \institution{Zhejiang University}
  % \streetaddress{1 Th{\o}rv{\"a}ld Circle}
  \city{Hangzhou}
  \country{China}
}
\email{yiyunzhou@zju.edu.cn}

\author{Qi Liu}
\affiliation{%
  \institution{Zhejiang University}
  % \streetaddress{1 Th{\o}rv{\"a}ld Circle}
  \city{Hangzhou}
  \country{China}
}
\email{qiliu@zju.edu.cn}

\author{Shulei Wang}
\affiliation{%
  \institution{Zhejiang University}
  % \streetaddress{1 Th{\o}rv{\"a}ld Circle}
  \city{Hangzhou}
  \country{China}
}
\email{shuleiwang@zju.edu.cn}

\author{Chang Yao}
\affiliation{%
  \institution{Zhejiang University}
  % \streetaddress{1 Th{\o}rv{\"a}ld Circle}
  \city{Hangzhou}
  \country{China}
}
\email{changy@zju.edu.cn}

\author{Jingyuan Chen${\dagger}$}
\affiliation{%
  \institution{Zhejiang University}
  % \streetaddress{1 Th{\o}rv{\"a}ld Circle}
  \city{Hangzhou}
  \country{China}
}
\email{jingyuanchen@zju.edu.cn}
% \author{
% \textbf{Wenkang Han},
% \textbf{Wang Lin},
% \textbf{Yiyun Zhou},
% \textbf{Qi Liu},
% \textbf{Shulei Wang},
% \textbf{Chang Yao},
% \textbf{Jingyuan Chen}\textsuperscript{*}\\
% }



% \keywords{Diffusion Model, Face Restoration, Identity Preservation}
%    <concept>
%        <concept_id>10010147.10010178</concept_id>
%        <concept_desc>Computing methodologies~Artificial intelligence</concept_desc>
%        <concept_significance>500</concept_significance>
%        </concept>
%  </ccs2012>

% \ccsdesc[500]{Computing methodologies~Artificial intelligence}



\renewcommand{\shortauthors}{Han and Lin, et al.}

\input{abstract}
\begin{teaserfigure}
\hsize=\textwidth
\centering
\includegraphics[width=0.95\textwidth]{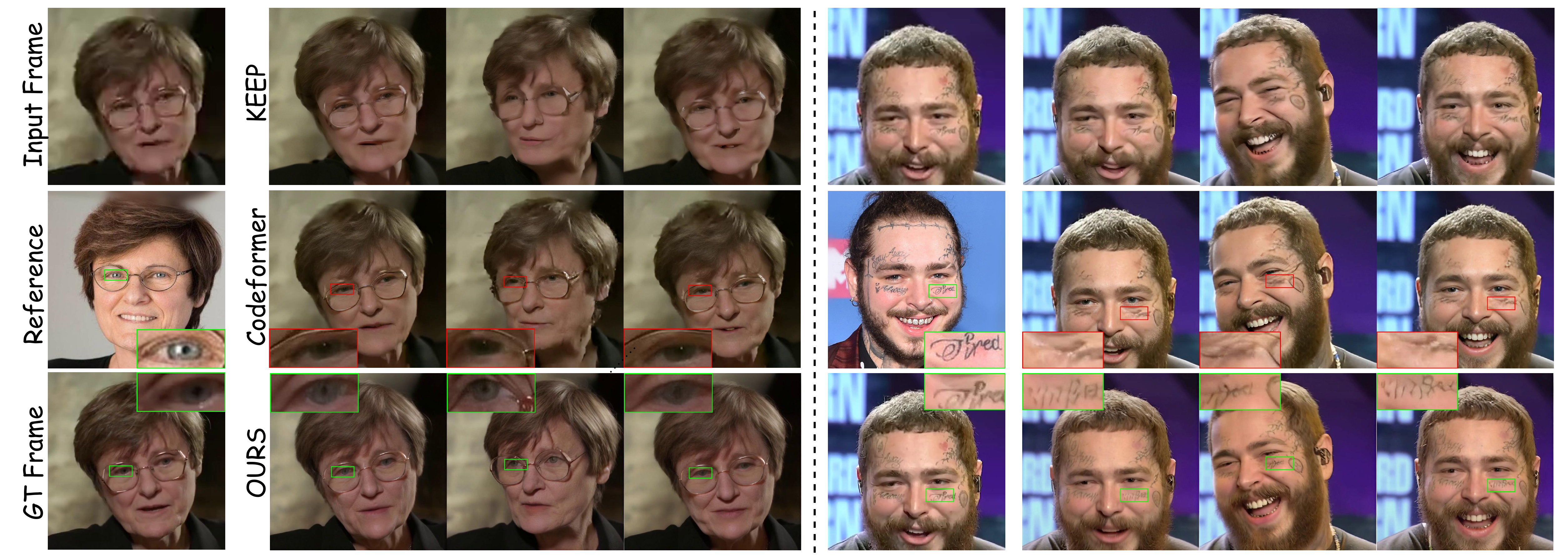}
\caption{Qualitative evaluation of the proposed IP-FVR method. In the left example, KEEP~\cite{feng2024kalman} exhibits substantial identity drift across frames, and Codeformer~\cite{zhou2022towards} fails to retain key identity details, such as light blue irises. In contrast, IP-FVR achieves superior visual quality and consistently preserves identity across frames. In the right example, baseline methods are unable to restore finer identity features, such as tattoos, whereas IP-FVR effectively retains and partially restores these finer details.}
% (\textit{Zoom in for a closer look}.)
\label{fig:main}
\end{teaserfigure}

\maketitle
\renewcommand{\thefootnote}{}
\footnotetext{$^{\dagger}$Corresponding authors.}

\input{intro}

\input{relative_work}

\input{method}
\input{experiment}

\section{Conclusion}
In this work, we introduced IP-FVR, a novel face video restoration method capable of recovering high-quality videos while preserving individual identities. By utilizing reference faces as visual prompt and incorporating identity information through decoupled cross-attention mechanisms, our approach generates detailed and identity-consistent results. Additionally, we introduce two key strategies to tackle identity drift: an identity-preserving feedback learning method that combines cosine similarity-based rewards with suffix-weighted temporal aggregation to minimize intra-clip drift, and an exponential blending strategy to address inter-clip drift by aligning identities across video segments. To address identity drift over extended sequences, we implemented an exponential blending strategy that maintains consistent identity representation and enhances temporal coherence. Experiments on both synthetic and real-world datasets demonstrate that IP-FVR outperforms existing methods in image quality and identity preservation. 
% The proposed method offers an effective solution for restoring degraded face videos, with significant potential for applications in video enhancement and editing.

% \newpage

\bibliographystyle{ACM-Reference-Format}
\bibliography{reference}

% \newpage
% \begin{appendices}
% \newpage
% \newpage
% \input{Section/Appendix}
% \end{appendices}

% \clearpage
\end{document}

%% file: abstract.tex
% \begin{abstract}
% Face Video Restoration (FVR) aims to recover high-quality face videos from degraded versions. Traditional methods struggle to preserve fine-grained, identity-specific features when degradation is severe, often producing average-looking faces that lack individual characteristics. To address these challenges, we introduce \textbf{IP-FVR}, a novel method that leverages a high-quality reference face image as a visual prompt to provide identity conditioning during the denoising process. \textbf{IP-FVR} incorporates semantically rich identity information from the reference image using decoupled cross-attention mechanisms, ensuring detailed and identity consistent results. To combat identity drift over extended video sequences, we propose an exponential blending strategy that combines latent codes from previous clips during each denoising iteration. This approach maintains a stable and consistent identity representation, effectively reducing frame-to-frame inconsistencies. Additionally, we enhance the restoration process with a multi-stream negative prompt, guiding the model's attention to relevant facial attributes and minimizing the generation of low-quality or incorrect features. Extensive experiments on both synthetic and real-world datasets demonstrate that IP-FVR outperforms existing methods in both quality and identity preservation, showcasing its substantial potential for practical applications in face video restoration. Our code and datasets are available at \textcolor{blue}{\url{https://ip-fvr.github.io/}}
% \end{abstract}

\begin{abstract}
Face Video Restoration (FVR) aims to recover high-quality face videos from degraded versions. Traditional methods struggle to preserve fine-grained, identity-specific features when degradation is severe, often producing average-looking faces that lack individual characteristics. To address these challenges, we introduce \textbf{IP-FVR}, a novel method that leverages a high-quality reference face image as a visual prompt to provide identity conditioning during the denoising process. \textbf{IP-FVR} incorporates semantically rich identity information from the reference image using decoupled cross-attention mechanisms, ensuring detailed and identity consistent results. For intra-clip identity drift (within 24 frames), we introduce an identity-preserving feedback learning method that combines cosine similarity-based reward signals with suffix-weighted temporal aggregation. This approach effectively minimizes drift within sequences of frames. For inter-clip identity drift, we develop an exponential blending strategy that aligns identities across clips by iteratively blending frames from previous clips during the denoising process. This method ensures consistent identity representation across different clips. Additionally, we enhance the restoration process with a multi-stream negative prompt, guiding the model's attention to relevant facial attributes and minimizing the generation of low-quality or incorrect features. Extensive experiments on both synthetic and real-world datasets demonstrate that IP-FVR outperforms existing methods in both quality and identity preservation, showcasing its substantial potential for practical applications in face video restoration. Our code and datasets are available at \url{https://ip-fvr.github.io/}.
\end{abstract}

%% file: intro.tex
\section{Introduction}
\input{Figure/degradation}
\textbf{Face Video Restoration} (FVR)~\cite{wang2021towards,gu2022vqfr,yang2021gan} aims to recover high-quality (HQ) face videos from diversely degraded versions, such as those affected by blur, downsampling, and random noise. The restoration process is inherently challenging due to the diverse and complex nature of these degradations, making it an ill-posed problem with multiple plausible solutions for a given low-quality (LQ) input. As shown in Figure~\ref{fig:degradation}, as degradation intensifies, the identity-specific features---such as eye bags, iris color, and nose contours---become progressively less distinguishable, leading to the failure of conventional restoration methods in preserving the fine-grained details that characterize individual identities. When we are familiar with a particular person's identity, it becomes easier to detect subtle differences in these details. 

% These subtle features are often easy to detect when we are familiar with a person's identity, but are harder to preserve as degradation worsens. 

%contextual identity information for specific individuals progressively diminishes. This often causes conventional face restoration methods to fail in accurately restoring identity-specific facial details, such as eye bags, iris color, and nose contours. When we are familiar with an individual’s unique facial characteristics, these subtle differences become more noticeable.

% As shown in Fig.~\ref{fig:degradation}, the dimensionality of the solution space increases significantly with the level of degradation, often resulting in outputs from general face restoration methods that fail to accurately capture identity-specific facial details (\textit{e.g.}, eye bags, pupil color, nose contour). When we are familiar with a particular person's identity, it becomes easier to detect subtle differences in these details. 

Recent advances in face restoration have leveraged generative priors~\cite{he2022gcfsr,yang2020hifacegan,yang2021gan,gu2020image,menon2020pulse,pan2021exploiting,pan2025generative,wang2025irbridge,yan2025diff}, pretrained codebook priors~\cite{gu2022vqfr,zhou2022towards,wang2022restoreformer} and diffusion priors~\cite{li2025set,zou2023flair,varanka2024pfstorer,lu2024robust,liang2024authface,lu2023tf,wang2025towards,lin2024non,wang2025discrete}, yielding impressive improvements in restoration quality. However, these methods often rely heavily on prior knowledge derived from extensive facial training datasets. When facial features degrade beyond the recognition capability of the model (\textit{e.g.}, Figure~\ref{fig:degradation}(d)), these methods tend to generate HQ images with averaged facial features. While realistic, these faces often fail to capture the critical identity-specific characteristics that distinguish one person from another.

%lose critical identity-specific characteristics. 
To address this limitation, some approaches~\cite{li2020enhanced,li2022learning,varanka2024pfstorer} introduce an HQ reference face image of the same identity to provide additional identity context, theoretically enhancing identity preservation. However, these methods face challenges due to variations in camera angle, expression, and lighting conditions between the reference and the LQ faces, often leading to rigid or unnatural expressions. Moreover, in the context of face video restoration, where the viewing angle continuously changes, the effective integration of identity information from a reference face image has not been fully explored.

In this paper, we tackle these challenges by introducing \textbf{IP-FVR}, a novel method that \textit{shows} the model the identity context through a reference face image and \textit{polishes} the restoration process to ensure high-quality, identity-preserving results. Specifically, IP-FVR extracts semantically rich, multimodal identity information from the reference face image and integrates this context into the restoration process using a decoupled cross-attention mechanism. This approach allows the model to focus on detailed identity-specific features, ensuring consistent and accurate restoration across frames.
The effectiveness of IP-FVR is qualitatively demonstrated in Figure~\ref{fig:main}, where it is compared with existing face video restoration methods, showing a superior balance between visual quality and identity preservation.
%Figure~\ref{fig:main} qualitatively compares the proposed method, IP-FVR, with existing approaches in face video restoration, highlighting its balance of quality and identity preservation. 

Furthermore, we are committed to addressing the challenge of identity drift. To tackle intra-clip (24 frames) identity drift, we introduce an identity-preserving feedback learning method that combines cosine similarity-based reward signals with suffix-weighted temporal aggregation. This approach effectively minimizes drift within sequences of frames. For inter-clip identity drift, we develop an exponential blending strategy that aligns identities across clips by iteratively blending frames from previous clips during the denoising process. This method ensures consistent identity representation across different video segments.

In summary, the main contributions of our study are as follows: \textbf{1)} We propose IP-FVR, a method that incorporates a reference face image as a visual prompt, independent of a text prompt, to provide conditioning information for identity customization in the denoising process. This approach achieves high-quality, identity-preserving face video restoration during inference. \textbf{2)} We propose an identity-preserving feedback learning method that combines cosine similarity-based rewards with suffix-weighted temporal aggregation to minimize intra-clip identity drift. \textbf{3)} We design an exponential blending strategy to address inter-clip identity drift by iteratively blending frames from previous clips during denoising, ensuring consistent identity representation across video segments. \textbf{4)} We conduct extensive experiments on both synthetic and real-world datasets, demonstrating a leading performance compared to previous methods and highlighting substantial potential for practical applications.

% \textbf{2)} We introduce an exponential blending strategy, which combines latent codes from previous clips in each denoising iteration, effectively mitigating identity drift in face video restoration. \textbf{3)} We conduct extensive experiments on both synthetic and real-world datasets, demonstrating a leading performance compared to previous methods and highlighting substantial potential for practical applications.

% 扩散模型在图像和视频的恢复和超分取得了令人印象深刻的表现。作为其中的一个子类，face restoration in image and video以正朝着diffusion model靠近。然而，这些方法要么如前面提到的一样，忽视在严重退化使用reference face image作为先验信息。要么只在图像粒度进行恢复预测，容易陷入prone to frame-to-frame inconsistencies。这种不一致性在face video restoration 领域具体表现为 video中人物的identity发生显著的波动 (see Fig XXX).

% However, as these methods are largely image-based, they are prone to frame-to-frame inconsistencies. In the field of facial restoration, this issue manifests as significant fluctuations in identity information across frames (see as Fig.[]), a phenomenon we refer to as \textit{identity drift}.

%% file: Figure/degradation.tex
\begin{figure}[t]
\centering
\includegraphics[width=0.47\textwidth]{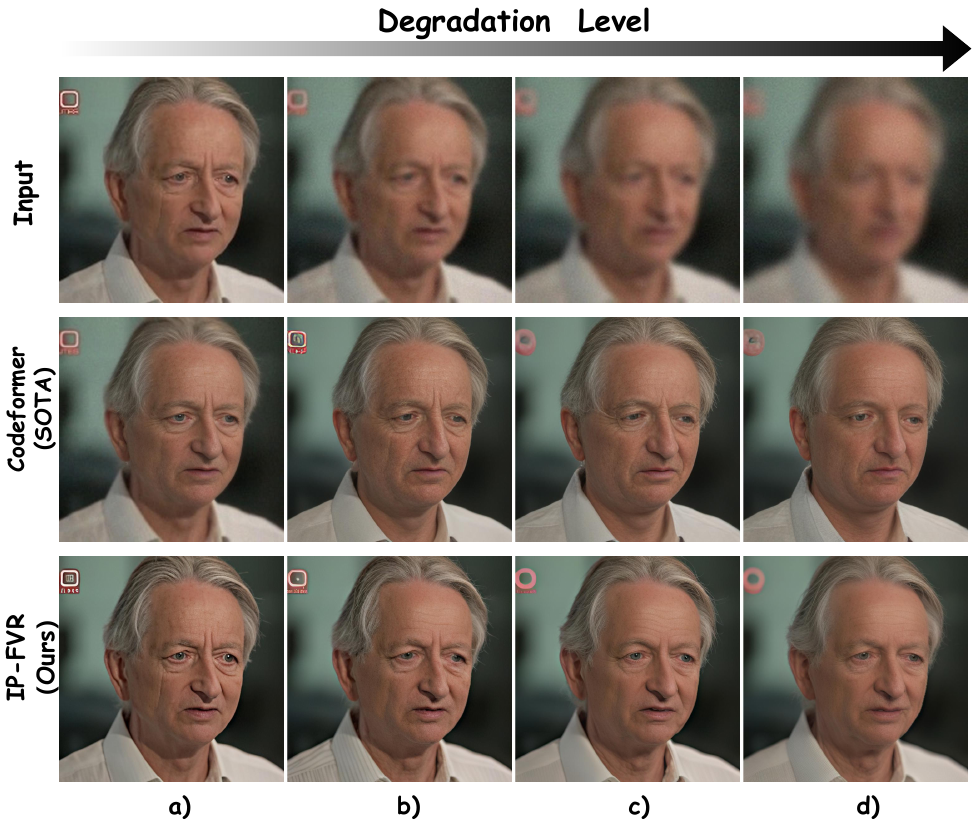}
\vspace{-1em}
% \caption{Comparison between IP-FVR and state-of-the-art (SOTA) method across varying levels of degradation. a) With minimal degradation, both the SOTA method and IP-FVR successfully preserve identity information. b) At moderate degradation, the SOTA method begins to inaccurately reconstruct facial contours. c) As degradation intensifies, the SOTA method loses more identity-specific details, introducing artifacts such as a double chin. d) Under severe degradation, the SOTA method produces an output with significant identity distortion, while the proposed IP-FVR continues to preserve key identity features, even with extremely low-quality inputs.}
\caption{Comparison of IP-FVR and the state-of-the-art (SOTA) method across degradation levels: a) At minimal degradation, both methods preserve identity well. b) With moderate degradation, the SOTA method begins to distort facial contours. c) As degradation increases, the SOTA method loses identity details and introduces artifacts like a double chin. d) Under severe degradation, the SOTA method shows substantial identity distortion, whereas IP-FVR maintains key identity features even at very low quality.}
\vspace{-2em}
\label{fig:degradation}
\end{figure}

%% file: relative_work.tex
\vspace{-1em}
\section{Related Works}
\label{related_work}
\textbf{Video Super-Resolution.} VSR aims to reconstruct high-resolution (HR) frames from degraded low-resolution (LR) video frames. Traditional VSR approaches~\cite{wang2019edvr,xue2019video,jo2018deep,isobe2020revisiting,9157178,feng20243,wu2024semantic,wang2024instruction,10.1007/978-3-030-58610-2_38,liang2022recurrent} typically rely on pre-defined degradation process~\cite{liu2013bayesian,nah2019ntire,xue2019video,yi2019progressive} (\textit{e.g.}, bicubic resizing, downsampling after Gaussian blur), which limits their generalizability in real-world settings. To enhance robustness, some recent works~\cite{chan2022investigating,xie2023mitigating} incorporate diverse degradation-based data augmentation. Nonetheless, CNN-based methods still face challenges in producing realistic textures due to limited generative priors. In contrast, recent methods such as DIFFIR2VR-ZERO~\cite{yeh2024diffir2vr}, MGLD\cite{yang2025motion}, UAV~\cite{zhou2024upscale}, and VEnhancer~\cite{he2024venhancer} leverage pre-trained generative diffusion models like Stable Diffusion~\cite{ho2020denoising,rombach2022high} to introduce strong diffusion priors, enabling more detailed and temporally consistent outputs for real-world VSR applications. While these techniques have demonstrated success in recovering rich texture details in general scenes, they continue to face challenges in effectively balancing the quality and fidelity of generated subjects. This is particularly true in facial scenarios, where even subtle alterations in facial features can jeopardize identity preservation.

\noindent\textbf{Face Restoration.} \textit{Generative prior-based} face restoration methods~\cite{he2022gcfsr,yang2020hifacegan,yang2021gan,gu2020image,menon2020pulse,pan2021exploiting} leverage pre-trained GANs like StyleGAN~\cite{karras2020analyzing} to enhance texture detail in degraded images. By projecting low-quality faces into the generator’s latent space, these methods treat restoration as conditional generation. Approaches such as GLEAN~\cite{chan2021glean} and GFPGAN~\cite{wang2021towards} further integrate priors into encoder-decoder structures, achieving a balance between fidelity and efficiency, though challenges remain under severe degradation. \textit{Codebook prior-based} methods, like VQFR~\cite{gu2022vqfr}, CodeFormer~\cite{zhou2022towards}, and RestoreFormer~\cite{wang2022restoreformer}, utilize pre-trained vector-quantized (VQ) codebooks as discrete dictionaries of facial features, achieving state-of-the-art performance in blind face restoration. Unlike continuous generative priors, these methods compress the latent space into a finite codebook, enhancing robustness to severe degradation. Through vector quantization and adversarial training, codebook priors effectively store high-quality facial details for improved restoration results. The latest advances~\cite{qiu2023diffbfr,zou2023flair,varanka2024pfstorer,liang2024authface,kuai2024towards,tao2024overcoming} employ \textit{diffusion priors}, harnessing their generative power to produce high-quality and robust face restorations. Despite the ability of these methods to generate detailed facial features, they often struggle to maintain identity fidelity when degradation is severe.

\noindent\textbf{Human Image Personalization.} In this paper, we primarily focus on preserving facial identity. Current human image personalization methods based on diffusion models mainly fall into two categories. The first, represented by approaches like FastComposer~\cite{xiao2024fastcomposer} and PhotoMaker~\cite{li2024photomaker}, encodes the reference image into one or more visual tokens, which are then fused with text tokens to serve as conditioning factors in the denoising process. The second, represented by works such as~\cite{ye2023ip,wang2024instantid,huang2024consistentid,han2025guirobotron,han2025contrastive,lu2024mace,gao2024eraseanything}, employs a decoupled cross-attention strategy that incorporates separate cross-attention layers specifically for the reference image. Although these strategies achieve high fidelity in identity preservation for text-to-image generation, applying these ideas to face video super-resolution while ensuring consistent identity across frames remains unexplored.

%% file: method.tex
\section{METHOD}
\input{Figure/train}
We propose a personalized face video restoration method, IP-FVR, that achieves both high texture detail and strong identity preservation. In Section \ref{metho:1}, we introduce the architecture of IP-FVR. As shown in Figure \ref{fig:overview}, this approach employs a face-to-text encoder and a face encoder to extract semantically rich identity information from a reference face. This identity information is then injected into the restored face video during the denoising process through decoupled cross-attention. In Section \ref{metho:2}, we present an identity-preserving feedback learning method that suppresses identity drift within a clip (24 frames) by combining cosine similarity-based reward signals with suffix-weighted temporal aggregation. To address identity drift across clips, we propose an exponential blending approach in Section \ref{method:3}. This approach aligns identities across clips by blending frames from previous clips during the iterative denoising process. Finally, to reduce the likelihood of generating low-quality restoration results or inaccurate face attributes, we propose a multi-stream negative prompting approach in Section \ref{method:4}.

% In Sec.~\ref{metho:1}, 我们介绍了如何在a short clip (24 frames)内进行高身份忠诚度的人脸视频恢复。如xxx图中所示，它使用XXX和visual Encoder从reference face中挖掘语义丰富的identity信息，并通过 decoupled cross-attention 在低质量人脸视频去噪过程中进行identity信息注入。为了支持在恢复长视频 (被切割成多个clips)减轻identity drift的现象，我们在Sec.X中介绍了exponential blending的方法，它通过在迭代去噪过程中混合过去clip的frame信息对齐clip之间的identity。

\subsection{Preliminaries}
\noindent\textbf{Personalized FVR Problems.} 
The face video restoration (FVR) aims to restore high-quality video $ hq \in \mathbb{R}^{F \times H \times W \times C}$ from low-quality inputs $lq  \in \mathbb{R}^{F \times h \times w \times C}$, where $F, H(h), W(w)$  and $C$ denote the video length, height, width, and channel, respectively. Letting $\mathcal{A}$ to represent the degradation process and $\mathcal{R}$ the restoration process, FVR can be denoted as $\hat{hq} = \mathcal{R}(lq)=\mathcal{R}(\mathcal{A}(hq,d))$, where $d$ represents a series of parameters in the degradation process (\textit{e.g.}, blur, downsampling, and random noise).  When the degradation parameter $d$ approaches infinity, the resulting images become nearly pure noise, making faithful restoration and identity preservation impossible. 
There exists a threshold $d_{th} < \infty$, beyond which faithful recovery is no longer achievable. However, if additional personalized priors $p_{id}$ are available, faithful restoration can still be achieved:
\begin{equation}
    \hat{hq} = \mathcal{R}(lq)=\mathcal{R}(\mathcal{A}(hq,d),p_{id}),
\end{equation}
as $p_{id}$ remains invariant with respect to any value of degradation $d$. In this paper, we incorporate this identity information in the noise prediction of the diffusion model.

\noindent\textbf{Video Latent Diffusion Model.} Our approach builds upon the pretrained video super-resolution architecture VEnhancer~\cite{he2024venhancer}, enabling high-detail video super-resolution. Given a pair of supervised training data ($lq,hq$). The $lq$ video is serves as input ($c_{lq}$) to ControlNet~\cite{zhang2023adding}, which conditions the denoising process of the Video Latent Diffusion Model (VLDM)~\cite{zhang2023i2vgen}. Next, pretrained variational autoencoder (VAE) encoder $\mathcal{E}$ compress $hq$ into a low-dimensional latent representation, denoted as $\boldsymbol{z} = \mathcal{E}(hq)$. 
while the corresponding decoder $\mathcal{D}$ maps the latent representation back to the pixel space, yielding $\hat{hq}=\mathcal{D}(\boldsymbol{z})$. 

In the diffusion process, noise is gradually added to the latent vector $\boldsymbol{z}$ over a total of $T$ steps. For each time step $t$, the diffusion process is represented as follows:
\begin{equation}
    \boldsymbol{z}_t =\alpha_t\boldsymbol{z} + \sigma_t\boldsymbol{\epsilon},   
\end{equation}
where $\alpha_t$ and $\sigma_t$ denote the noise schedule parameters, with the corresponding log signal-to-noise ratio ($\mathit{i.e.}$, $log(\alpha_t^2 / \sigma_t^2)$), monotonically decreasing as $t$ increases. In the denoising stage, By adopting v-prediction parameterization~\cite{salimans2022progressive}, the U-Net denoiser model $f_{\theta}$ learns to make of predictions of $\boldsymbol{v}_t = \alpha_t \boldsymbol{\epsilon} - \sigma_t \boldsymbol{z}$. It receives the diffused latent $\boldsymbol{z_t}$ as input and is optimized by minimizing the denoising score matching objective:
\begin{equation}
    \label{eq:vldm}
    \mathcal{L}_{\text{rec}}=
     \mathbb{E}_{\boldsymbol{z}, c_{\text{text}},c_{lq}, \epsilon \sim \mathcal{N}(0, \mathbb{I}), t} \left[ \left\| \boldsymbol{v} - f_\theta (\boldsymbol{z_t}, t, c_{\text{text}},c_{lq}) \right\|_2^2 \right].
\end{equation}

\subsection{IP-FVR Architecture}
\label{metho:1}
% especially under severe degradation,
Existing FVR methods struggle to faithfully restore identity-consistent facial videos, primarily due to the lack of stable identity information. To address this, we propose to incorporate prior information from a personalized reference face image for face video restoration, aiming to preserve identity consistently across frames. As shown in Figure~\ref{fig:overview}, proposed IP-FVR utilize a face encoder and a face2text encoder to extract rich identity information from the reference face into text and image prompts to guide the denoising process in U-Net. Specifically, the face2text encoder first uses a face attribute detector to identify identity-related keywords, which are then transformed into identity-specific text features via an LLM and CLIP text encoder. Furthermore, to achieve restoration results that more accurately capture each identity, we independently train corresponding LoRA weights for each identity.

% 如XXX图所示，我们分别使用visual encoder和[xx 管道]从reference face 中提取丰富的identity 信息作为text 和 image prompt 指导U-Net的去噪过程。其中[xx管道]首先使用Face attribute detector得到identity关键词，然后通过LLM 和CLIP-encoder转为identity相关的文本特征。此外，to achieve restoration results that better capture each identity
% we independently train the corresponding LoRA weights for each identity.
% Finally, we independently train the corresponding LoRA weights for each identity to align the pre-trained decoupled cross-attention weights with the frozen base model. 
% % and achieve restoration results that better capture each identity, 

\noindent\textbf{Decoupled Cross-Attention.} Inspired by the recent success of customized image generation~\cite{ye2023ip,wang2024instantid,huang2024consistentid}
% 感觉还是不要出现具体的方法名吧，再几一两个ip后续的延伸工作
in text-to-image generation---where decoupled cross-attention enables fine-grained control over image features while preserving text-prompt compatibility---we extend this concept to video face restoration. 
The decoupled cross-attention requires two inputs: an image prompt $c_i$ and a text prompt $c_t$. $c_i$ is obtained from the reference face via a visual encoder, 
while $c_t$ is derived from the reference face through a face2text encoder. The face2text encoder extracts identity-specific text features from the reference face using a face attribute detector and text encoders. Given the query features $Z$, which is the output of the U-Net block, we integrated decoupled cross-attention from~\cite{ye2023ip} into the 2D framework of the vanilla VLDM as follows:

\begin{equation}
    Z'=\text{Attention}(Q, K^t, V^t) + \lambda \cdot \text{Attention}(Q, K^i, V^i), 
\end{equation}
where $Q$, $K^t$, $V^t$ are the query, key, and value matrices of the attention operation for text cross-attention, $K^i$ and $V^i$ are for image cross-attention. Specifically, matrices $Q = Z W_q$, $K^i = c_i W_k^i$, $ V^i = c_i W_v^i$, $K^t = c_t W_k^t$, $V^t = c_t V_k^t$.
% enabling the injection of reference face features into each frame. 

\noindent\textbf{Face2Text Encoder.}
In the denoising process of the U-Net, text prompts play a crucial role in controlling facial features, expressions, and actions. 
% 一个好的text prompts（应该/包含）在精准的描述面部的特征的基础上，同时捕捉面部的神态。
% 对于精准的面部特征描述，我们XX
% 对于面部神态描述，我们XX
% 最后我们得到了什么什么，可以用来xx
% 多模态大模型的能力
For personalized FVR, the text prompt should not only accurately describe the facial features but also capture the subject's expression.
To achieve this, we use the facial attribute detector model~\cite{rudd2016moon} to extract the 40 facial attributes defined by CelebA~\cite{liu2015faceattributes}, generating a detailed list of facial attributes. Additionally, we use real-time face detection and emotion classification model~\cite{arriaga2017real} to annotate the emotion of video clips. Finally, we input the extracted facial attributes, emotion types, and manually labeled facial actions (\textit{e.g.}, speaking, smiling) as individual keywords into a large language model, which organizes them into natural language descriptions to enhance compatibility with the CLIP text encoder.

\noindent\textbf{Personalized LORA Fine-Tuning.}
The identity preservation benefits of directly integrating the pre-trained decoupled cross-attention module into a frozen vanilla VLDM vary significantly across different individuals. To better adapt to each identity, we adopt a LoRA-based fine-tuning approach. By leveraging Low-Rank Adaptation (LoRA)~\cite{hu2021lora}, we can efficiently adapt the model with minimal additional parameters, making it feasible to perform few-shot fine-tuning on a specific identity. After fine-tuning for each identity, the corresponding LoRA parameters $\psi_{id}$ are stored and can be directly applied during inference in a plug-and-play manner. As shown in Figure~\ref{fig:overview}, we fine-tune the spatial and temporal layers of the vanilla VLDM using trainable LoRA linear layers. This approach enables end-to-end few-shot training, ensuring consistent alignment of the reference face identity across a 24-frame clip. 

\input{Figure/reward}
\subsection{Identity Preserving Feedback Learning}
\label{metho:2}
In the domain of face video restoration, preserving identity consistency stands as a paramount objective.  However, existing methods often encounter the issue of identity drift,  where the generated video frames exhibit inconsistent identity characteristics across different time segments. To address this challenge, we introduce an identity-preserving mechanism based on feedback learning.

As shown in Figure~\ref{fig:reward}, we define a reward signal $R_f$ to measure the identity similarity between the generated frame $x_f$ and the reference face $x_{ref}$. The reward signal is computed by extracting facial feature vectors using a face detector and a face encoder, followed by calculating the cosine similarity between the generated frame and the reference face:
\begin{equation}
 r_f=\text{cos\_sim}(x_f,x_{ref}),
\end{equation}
where $x_f$ is the feature vector of the generated frame, and $x_{ref}$ is the feature vector of the reference face. To enhance the stability of the reward signal, we normalize $r_f$ as follows:
\begin{equation}
     A_f=\frac{r_f-\text{Mean}(\{r_0,r_1,...,r_F\})}{\text{Std}(\{r_0,r_1,...,r_F\})}.
\end{equation}
where Mean and Std denote the mean and standard deviation, respectively. 

In video diffusion models, the restoration of each frame inherently influences subsequent frames due to the model's reliance on temporal dependencies to maintain coherence across the video sequence. A deviation in one frame's restoration can propagate through the sequence, potentially leading to identity drift in later frames. To address this challenge, we employ a suffix-weighted reward mechanism to get final reward signal $R_f$, which assigns higher weights to more recent frames:
\begin{equation}
    R_f=\sum_{f'=f}^{F}\gamma^{f'-f}A_f,
\end{equation}
where $\gamma$ is a discount factor used to balance the weights of rewards at different time steps. 
To optimize identity preservation, we define a loss function based on the reward signal:
\begin{equation}
     \mathcal{L}_{\text{id\_reward}}=
     \mathbb{E}_{c_{\text{text}},c_{lq},c_{\text{img}}} \left[ 1-\text{exp}(\text{random$\{R_0,R_1,...,R_F\}$}) \right].
\end{equation}
This loss function maximizes the reward signal $R_f$, ensuring that the generated video frames maintain identity consistency with the reference face.

\subsection{Identity Stability in Long FVR}
\label{method:3}
To restore face videos of arbitrary duration, We employ a divide-and-conquer strategy by segmenting the video into clips, each consisting of 24 frames. This clip is processed independently in a single inference pass, and the outputs are concatenated for the final result. However, this straightforward approach of restoring each clip independently leads to significant temporal inconsistencies, which we call identity drift (see Figure~\ref{fig:cross_frame}). To address this issue, we propose an approach for inter-clip identity stabilization.

\noindent\textbf{Noise Sharing.} First, an intuitive improvement is to introduce overlapping frames between clips and utilize shared noise to enhance consistency. Specifically, for a given long low-resolution face video, we first divide it into $n$ overlapping clips $\mathcal{V}_1,\mathcal{V}_2,\ldots,\mathcal{V}_n$, each with clip length $F=24$ and overlap length $O=8$. For clip $\mathcal{V}_1$, we sample noise $\boldsymbol{\epsilon_1} \sim \mathcal{N}(0, I)$, where $\boldsymbol{\epsilon_1} \in \mathbb{R}^{F \times H \times W \times C}$. For $\boldsymbol{\epsilon_i}$ ($i>1$), it shares noise with the preceding clip $i-1$, which can be formulated as:
\begin{equation}
    \boldsymbol{\epsilon_{i}}=\text{stack}([\boldsymbol{\epsilon}_{i-1}^{F-O+1:F},\hat{\boldsymbol{\epsilon_i}}]),
\end{equation}
where $\hat{\boldsymbol{\epsilon_i}} \sim \mathcal{N}(0, I)$ and $\hat{\boldsymbol{\epsilon_i}} \in \mathbb{R}^{(F-O) \times H \times W \times C}$. 

While this approach helps mitigate identity drift within smaller regions, the issue still persists when comparing clips over a longer span. To further mitigate identity drift, we propose an \textbf{exponential blending} approach in addition to noise sharing. Specifically, denote $z_{t}(i)$ as the latent encoding of $\mathcal{V}_i$ at denoising step $t$. We perform exponential blending in the latent space, which can be formulated as:
\begin{equation}
 z_t^{1:O-1}(i)=\text{stack}([z_t^{F-2^j+1:F}(i-s+j)]_{j=0}^{s-1}),
\end{equation}
where $2^s=O$, so $s=3$ in this case. We recursively apply the above operations during the denoising steps in the super-resolution process of clip $\mathcal{V}_i$, recording intermediate results throughout. Compared to merely sharing noise, exponential blending mitigates the identity drift issue over a broader range by blending the latent encodings of multiple preceding clips across several denoising steps.

\input{Figure/quali_1}
\subsection{Inference with Negative Prompt}
\label{method:4}
In the inference stage, accurately controlling the restored face video to closely match the desired conditions presents a notable issue. Classifier-Free Guidance (CFG)~\cite{ho2022classifier} introduced a strategy combining conditional and unconditional descriptions to guide model generation. Inspired by this approach, most diffusion models now incorporate negative prompts to suppress low-quality image generation and enhance detail reconstruction accuracy. Our base model~\cite{he2024venhancer} leverages a positive quality prompt $pq$ (\textit{e.g.}, \textit{``Cinematic, High Contrast, Highly Detailed.''}) and a negative quality prompt $nq$ (\textit{e.g.},\textit{``painting, oil painting, sketch.''}). 
In combination with the text prompt $c_t$ and image prompt $c_i$ mentioned in Section~\ref{metho:1}, we propose a \textbf{multi-stream negative prompt} to guide the model away from non-existent facial attributes and achieving results with high detail and strong identity preservation.
During each denoising step, we integrate the outputs generated by these various prompts to obtain the final output:
\begin{equation}
    \tilde{z}_{t-1} = (1+w_{nt}+w_{nv})z^{pos}_{t-1} - w_{nt}z^{nt}_{t-1} - w_{nv} z^{nv}_{t-1},
\end{equation}
where $w_{nt}$ and $w_{nv}$ are the hyperparameters, and
\begin{equation}
\begin{split}
   z^{pos}_{t-1}=\epsilon_{\theta}(z_t,t,c_{lq},c_{pt \oplus pq},c_i) \\
   z^{nt}_{t-1}=\epsilon_{\theta}(z_t,t,c_{lq},c_{nt \oplus nq},c_i) \\
   z^{nv}_{t-1}=\epsilon_{\theta}(z_t,t,c_{lq},c_{pt \oplus pq},\tilde{c_i}),
\end{split}
\end{equation}
$pt$ is the positive text prompt as shown in Figure~\ref{fig:overview}, and $nt$ is the negative text prompt containing false facial attributes. $c_i$ and $\tilde{c_i}$ denote the facial visual features encoded by the visual encoder from the high-quality reference face and the degraded reference face, respectively.

%% file: Figure/train.tex
\begin{figure*}[t]
\centering
\includegraphics[width=0.90\textwidth]{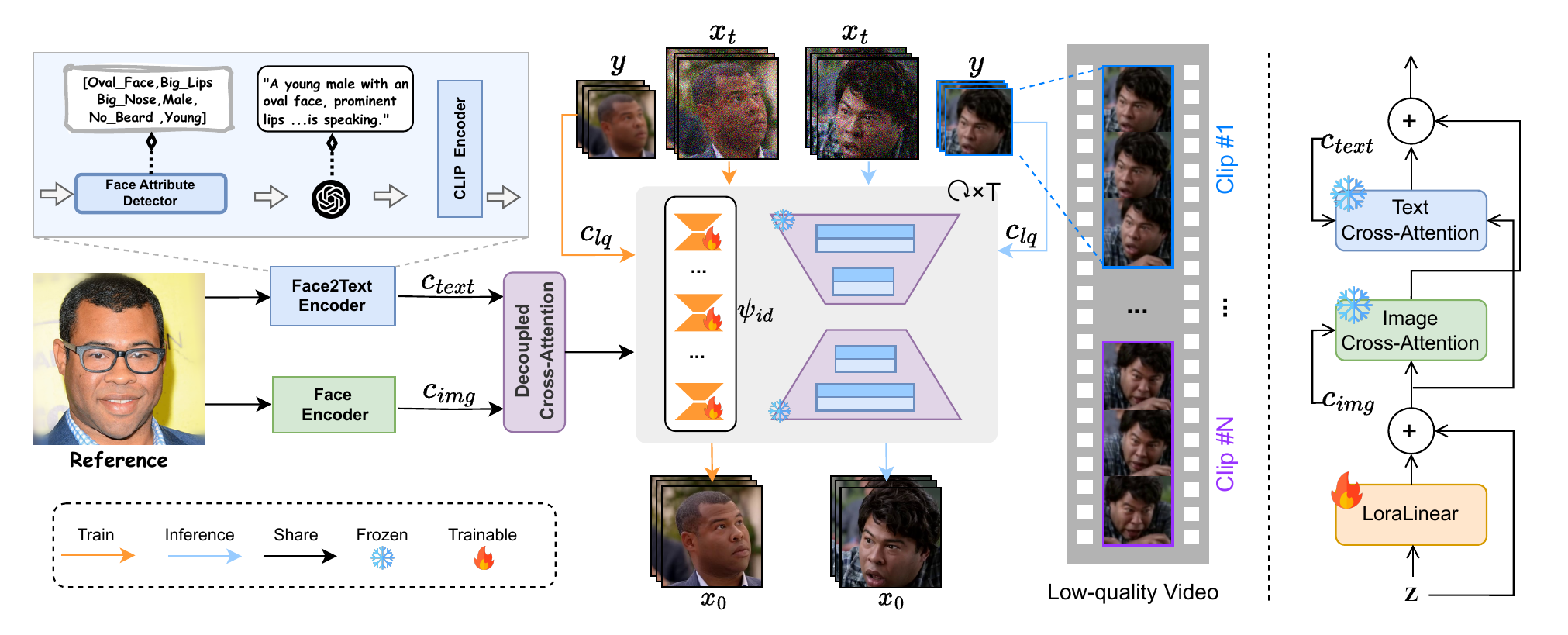}
\vspace{-1em}
\caption{The left diagram presents an overview of the fine-tuning and inference process of IP-FVR. It extracts multimodal features related to identity from the reference face using a face2text encoder and a face encoder. These features are then injected into the denoising process of the U-Net through decoupled cross-attention, enabling the restoration of identity-consistent face videos. The right diagram illustrates the network structure of the decoupled cross-attention mechanism.}
\vspace{-1em}
\label{fig:overview}
\end{figure*}

%% file: Figure/reward.tex
\begin{figure*}[t]
\centering
\includegraphics[width=0.9\textwidth]{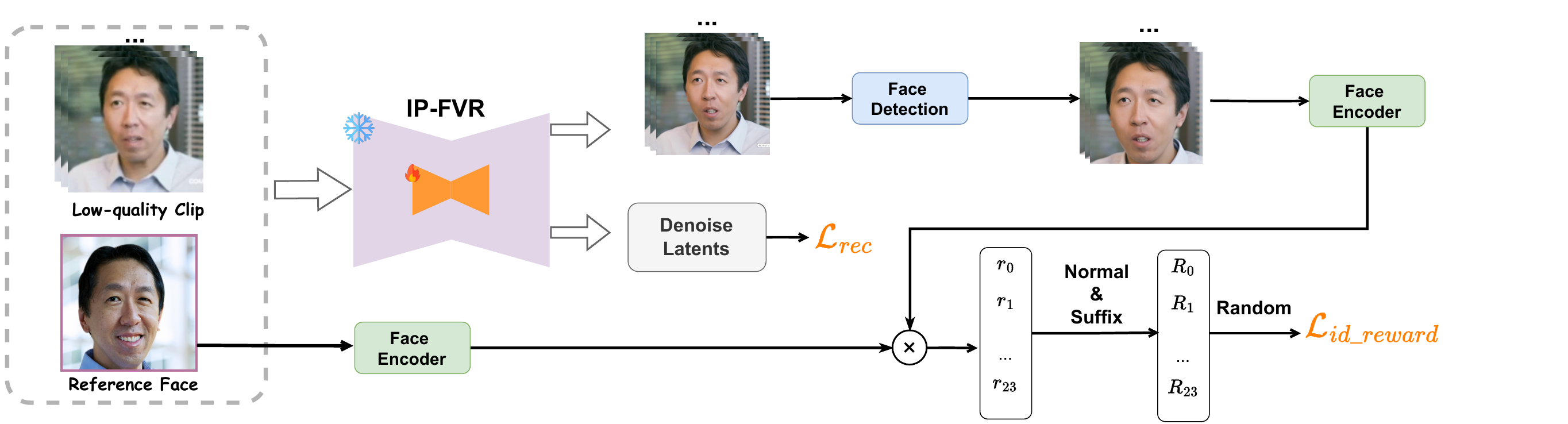}
\vspace{-1em}
\caption{Training process of the proposed IP-FVR. Combining Video Diffusion Model noise prediction loss $\mathcal{L}_{\text{rec}}$ and identity preservation loss $\mathcal{L}_{\text{id\_reward}}$  for Training.}
\vspace{-1em}
\label{fig:reward}
\end{figure*}

%% file: Figure/quali_1.tex
\begin{figure*}[t]
\centering
\includegraphics[width=0.9\textwidth]{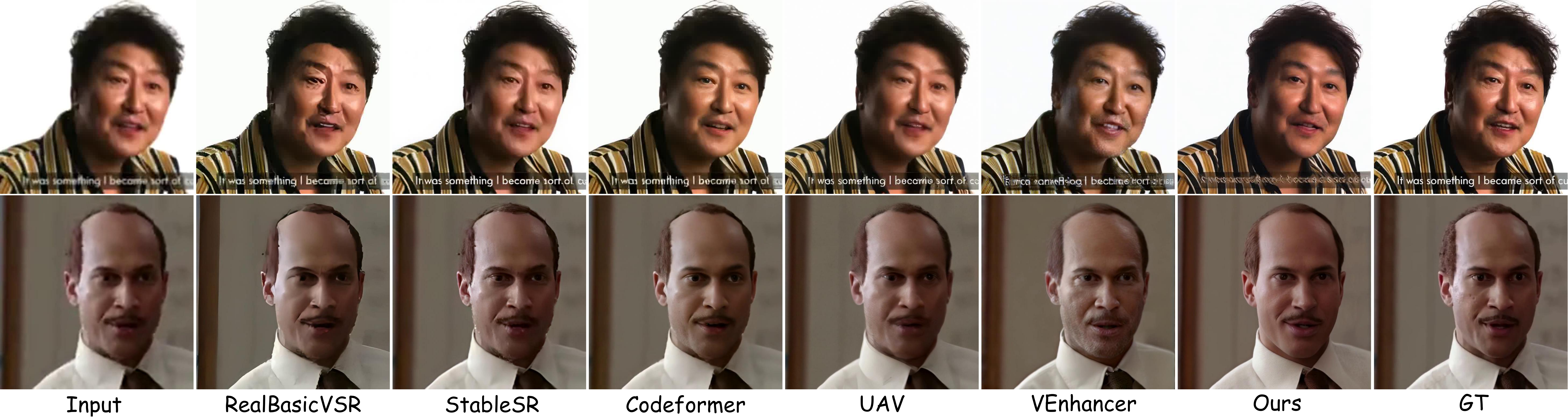}
\vspace{-1em}
\caption{Qualitative comparison on YouRef-light. IP-FVR produces higher restoration quality while maintaining high fidelity.}
\vspace{-0.5em}
\label{fig:quali_1}
\end{figure*}

%% file: experiment.tex
\input{Figure/quali_2}
\input{Figure/quali_3}
\section{Experiments}

\noindent\textbf{Datasets.}
To evaluate our proposed method's performance under both synthetic and real-world degradation scenarios, we utilized two datasets: (1) YouRef dataset, created by collecting high-quality videos of 18 celebrities from YouTube\footnote{https://www.youtube.com/}, with face regions extracted at 720x720 resolution. Corresponding reference face images were sourced from Google\footnote{https://www.google.com/} and Bing\footnote{https://www.bing.com/}. This dataset includes two variants with light and heavy degradation settings, alongside ground-truth data for synthetic degradation scenario evaluation. (2) FOS-V~\cite{chen2024towards} dataset, which features heterogeneous real-world scenarios, including interviews, sports, nature footage, and vintage films. For personalized video face super-resolution, we filtered this dataset to obtain facial clips of 20 celebrities, each paired with a corresponding reference face image.

\noindent\textbf{Implementation.}
Our personalized face video restoration method customizes an individual's identity information by using low- and high-quality video pairs ($lq, hq)$ from four different scenes, and performs 100 steps on a single A800 GPU. We ensure that the scenes used for personalized LORA fine-tuning do not appear in the test set. For other baseline methods, we perform inference on the same test set used by the proposed method, with parameters consistent with those in the corresponding papers. The light degradation involves applying first-order degradation, formulated as:
\begin{equation}
\label{eq:degradation}
     \mathbf{lq}= \left[\left(\mathbf{hq} \otimes \mathbf{k}_\sigma + \mathbf{n}_\delta \right) \downarrow_{r}\right]_{\text{FFMPEG}}.
\end{equation}
Here $\mathbf{k}$, $\mathbf{n}$ and $r$ represent the blur kernel, additive noise, and downsampling factor, respectively, with $r$ fixed at 4. We use a constant rate factor ($crf$) to control the degree of compression applied by FFMPEG. $crf$ adjusts the bitrate automatically to achieve a specified level of quality. The sampling intervals for $\sigma$, $\delta$ and $crf$ are $[0.2,3]$, $[1,5]$ and $[18,35]$, respectively. 

The heavy degradation involves the application of second-order degradation. Specifically, we employ Equation~\ref{eq:degradation}, with $\sigma$, $\delta$, and $crf$ sampled from the ranges $[2,5]$, $[1,10]$, and $[18,35]$, respectively. During the first round of degradation, the downsampling factor $r$ is fixed at 4, whereas in the second round (if applied), $r$ is set to 1. Additionally, there is a 90\% probability of performing a second round of degradation.

\noindent\textbf{Evaluation Metrics.}
For the synthetic dataset with ground truth, we evaluate performance using PSNR, SSIM, and LPIPS~\cite{zhang2018unreasonable}. For the real-world dataset, we employ CLIP-IQA~\cite{wang2023exploring}, MUSIQ~\cite{ke2021musiq}, and LIQE~\cite{zhang2023blind}. Additionally, we assess identity preservation using IDS (cosine similarity with ArcFace~\cite{deng2019arcface}). To measure inter-frame consistency, we utilize flow warping error $E_{warp}$~\cite{lai2018learning} and $\sigma_{IDS}$, where $\sigma_{IDS}$ represents the standard deviation of identity similarity across the entire face video.

\subsection{Comparison with State-of-the-Art}

\noindent\textbf{Qualitative Evaluation.} To evaluate the effectiveness of the proposed method, we present visual comparisons of single-image results from Figure~\ref{fig:quali_1} to Figure~\ref{fig:quali_3}. As shown in Figure~\ref{fig:quali_1}, Codeformer~\cite{zhou2022towards} and VEnhancer~\cite{he2024venhancer} achieve the highest quality results among the baseline methods. However, they also introduce artifacts that alter identity characteristics, such as the addition of unintended facial hair or the transformation of single eyelids into double eyelids. In contrast, the proposed method remains faithful to the identity and is able to recover low-level identity features, such as skin texture. Figure~\ref{fig:quali_2} showcases examples with severe degradation, where other methods produce distorted facial features or lose critical identity information. Our method, however, retains both quality and identity consistency. Figure~\ref{fig:quali_3} compares results on a real-world FOS-V dataset, demonstrating that our approach generates facial features that align with the reference identity, such as \textit{Pacino's apple-shaped chin} and \textit{Beyoncé's brown eyes}. 
Finally, on the right side of Figure~\ref{fig:cross_frame}, we present a comparison between VEnhancer and our method on a representative example, spanning 20 frames from the start to the end. The experimental results demonstrate that our method achieves higher identity similarity and effectively reduces identity drift.
\input{Table/main_exp}
\input{Table/rw_exp}

% 为了评估所提出的方法的有效性。我们在图x到图y展示了进行了单图的视觉比较。从Fig 4我们可以看出，codeformer和venhancer在所有基线中取得了最好的质量效果，但是也出现了虚假的identity信息，比如多余的胡须，单眼皮变成双眼皮，而所提出的方法忠于identity并能恢复皮肤纹理这类底层identity信息。Fig 5展示了具有严重退化的示例，可以发现其他的方法出现了扭曲的五官或严重失去identity 信息，而所提出的方法依旧能兼顾质量和identity 保留。Fig 6展示在real world的FOS-V上的比较，所提出方法能够根据reference先验，生成符合identity的面部特征，如pacino的
% apple-shaped chin和Beyoncé的棕色眼睛。此外，我们在图X的右侧展示VEnhancer和我们方法在一个具有代表性的例子下跨度为20帧下效果对比，结果表明我们的方法具有更高的identity similarity 和更低的identity drift。

\noindent\textbf{Quantitative Evaluation.}
The quantitative results on the synthetic YouRef-heavy test set are presented in Table~\ref{syn_data}. Through identity preserving feedback learning, IP-FVR effectively preserves and incorporates identity-specific information, facilitating efficient identity integration during the inference stage. This approach achieves superior performance in PSNR, SSIM, and identity-preserving metric IDS. Additionally, IP-FVR maintains strong temporal consistency, as evidenced by superior performance on metrics $E_{warp}$ and $\sigma_{IDS}$, indicating smoother scene transitions and reduced identity drift across frames. Another noteworthy finding is that diffusion-based methods, such as UAV~\cite{zhou2024upscale} and VEnhancer~\cite{he2024venhancer}, excel in no-GT image quality metrics CLIP-IQA, MUSIQ, and LIQE. This suggests that, in comparison to traditional methods, diffusion model-based methods are capable of generating richer texture details, thereby enhancing overall visual quality.

Moreover, as shown in Table~\ref{rw_data}, the proposed method achieves the highest single-image quality on the FOS-V dataset, owing in part to the diffusion model's prior knowledge, which enables the generation of high-realism, high-detail images. Moreover, the multi-stream negative prompt approach we propose further guides the model to generate outputs that align with both the text and image prompt descriptions, contributing to this improved performance.

\input{Figure/ab-1}

\input{Figure/ab-2}
\subsection{Ablation Study}
\noindent\textbf{Effectiveness of Decoupled Cross-Attention.}
% In Table~\ref{tab:ab1}, we present the performance of IP-FVR under conditions where trainable LoRA weights (\textbf{w/o.FT}) and the reference face image (\textbf{w/o.REF}) are removed, respectively. Results indicate that using the face prompt enhances the identity consistency metric IDS, and through few-shot training, the PSNR, SSIM and LPIPS metrics further improve.
Table~\ref{tab:ab1} presents the performance of IP-FVR with different prompt configurations. In this context, removing the Face2Text encoder indicates that the face attribute detector is omitted for identity-specific facial feature extraction, with a simple text prompt, such as \textit{``a good video''}, used as a substitute. Meanwhile, removing the visual encoder signifies that only the text modal is utilized for cross-attention. The results show that the model's performance is negatively impacted by the removal of either the Face2Text encoder or the Visual encoder, with a particularly notable decrease in the IDS metric by 12.7\% and 6.3\%, respectively. This suggests that both the identity-specific facial attributes from the text modality and the facial features from the visual modality provide meaningful guidance for identity preservation in face video restoration.

\noindent\textbf{Effectiveness of Identity Preserving Feedback Learning.}
Additionally, Figure~\ref{fig:ab-1} presents comparative examples before and after applying the identity preserving feedback learning. Directly using a pluggable decoupled cross-attention weights~\cite{ye2023ip} enables the injection of naive identity attributes (\textit{e.g.}, \textit{Chen’s nose shape and Katalin’s eye color}) into the output. However, this approach can also result in fixed expressions, such as the \textit{unnatural openness of Chen’s eyes}. In contrast, the complete model better aligns subtle identity characteristics, such as \textit{skin texture}, producing results with high detail and high identity preservation.

\noindent\textbf{Effectiveness of Exponential Blending.}
Table~\ref{tab:ab2} presents a comparison of temporal consistency metrics for IP-FVR across various configurations. When the Exponential Blending approach is removed (\textit{i.e.}, \textbf{w/o EB}), the model’s $\sigma_{IDS}$ and $E_{warp}$ scores increase by 0.432 ($\times 10^{-2}$) and 0.549 ($\times 10^{-3}$), respectively. Removal of Noise Sharing (NS) results in additional increases in IP-FVR's $\sigma_{IDS}$ and $E_{warp}$. This result highlights the importance of utilizing exponential blending during inference to produce face video restoration outputs with smooth transitions and stable identity features. 

% Additionally, we observed that even the \textbf{w/o NS\&EB} version of IP-FVR, achieves lower $\sigma_{IDS}$ compared to other diffusion-based methods. This improvement is attributed to the reference face serving as an identity prior, acting as an anchor across frames within a clip to guide each frame’s identity features to align with it.

Figure~\ref{fig:cross_frame} illustrates the identity similarity across frames for a representative face video. As shown in the left panel, our method achieves the highest average identity similarity with minimal drift. IP-FVR significantly reduces frame-to-frame identity fluctuations compared to VEnhancer, exhibiting only a 5\% drift between frames 35 and 55 versus VEnhancer's 14.4\%. Diffusion-based methods often suffer higher fluctuations because computational constraints necessitate processing videos in separate clips without temporal attention. Our exponential blending approach effectively addresses this issue.

\input{Table/tab-ab1}

\input{Table/tab-ab2}

\input{Table/tab-ab3}

\noindent\textbf{Effectiveness of Negative Prompt.}
We conducted an ablation study on negative prompts using the YouRef-heavy dataset, with the results presented in Table~\ref{tab:ab3}. It indicates that combining the text prompt and image prompt—generated by the face2text encoder and visual encoder shown in Figure~\ref{fig:overview}—with the corresponding negative prompts for text and visual modalities (see Section~\ref{method:3}) achieves optimal LPIPS and IDS scores. This demonstrates that the proposed multi-stream negative prompt effectively guides the model away from non-existent facial features, thus enabling high identity-preserving face restoration. Additionally, we explored the hyperparameters of the negative text prompt ($w_{nt}$) and negative visual prompt ($w_{nv}$), with results indicating that the optimal configuration is $w_{nt}=0.5$ and $w_{nt}=0.5$.

% \subsection{User Study}
% Quantitative metrics often struggle to fully capture the subtle nuances of human preferences in perceived quality. To address this limitation, we conducted a user study to evaluate four methods: diffusion-based image and video super-resolution methods (StableSR~\cite{wang2024exploiting} and UAV~\cite{zhou2024upscale}), a state-of-the-art face restoration method (Codeformer~\cite{zhou2022towards}), and our proposed approach, IP-FVR.

% We recruited 20 participants for the user study. Each participant evaluated 10 sets of anonymized videos: 5 sets were randomly selected from the restoration results of the four methods on YouRef-light along with their corresponding ground truth, and the remaining 5 sets were randomly selected from YouRef-heavy. For each set, participants were tasked with selecting the index of the video with the highest visual quality and the index of the video with the highest identity fidelity. As shown in Figure~\ref{fig:user_study}, the results of the user study demonstrate a clear preference for our method over others in terms of both visual quality and identity fidelity.

% \begin{figure}[h]
% \centering
% \includegraphics[width=0.47\textwidth]{Fig/user_study_results.pdf}
% \vspace{-1em}
% \caption{ User study result.}
% \vspace{-0.5em}
% \label{fig:user_study}
% \end{figure}

% \subsection{User Study}
% 如果工作量大可以放进正文，否则放到附录水一下

% \input{Figure/prompt}
% \input{Figure/time_consistency}

%% file: Figure/quali_2.tex
\begin{figure}[h]
\centering
\includegraphics[width=0.48\textwidth]{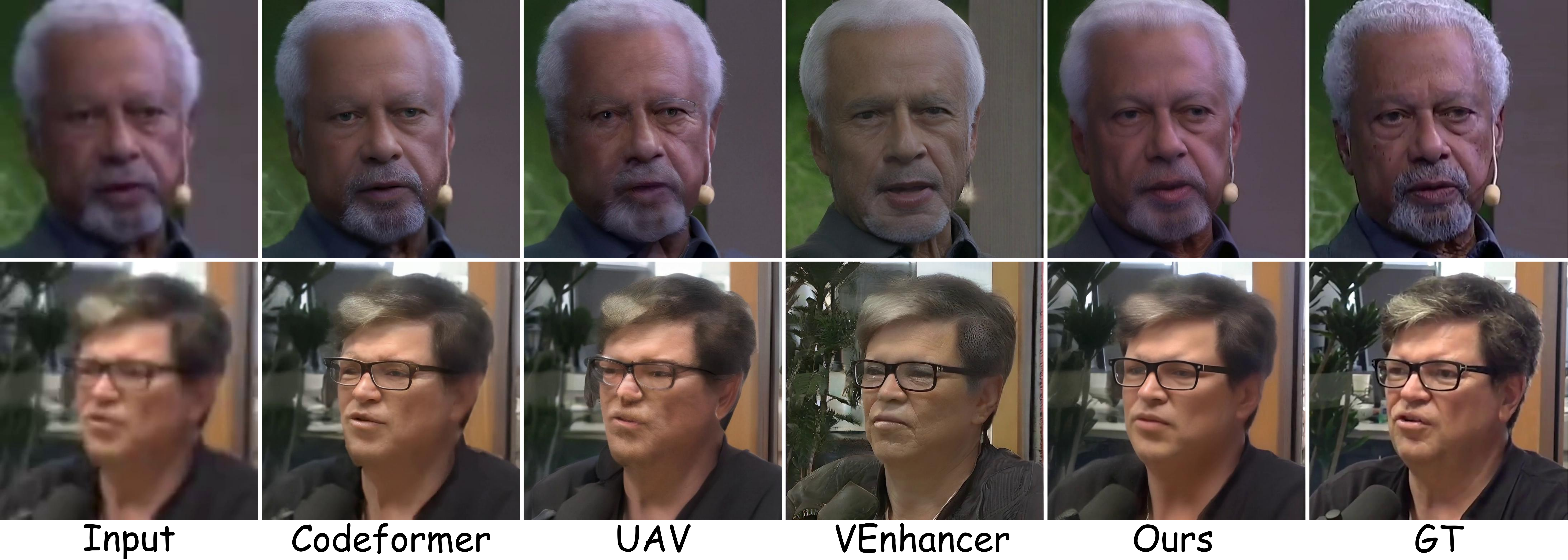}
\vspace{-2em}
\caption{Qualitative comparison on YouRef-heavy.}
% Even under severe degradation, IP-FVR effectively preserves facial identity.}
\vspace{-1em}
\label{fig:quali_2}
\end{figure}

%% file: Figure/quali_3.tex
\begin{figure*}[t]
\centering
\includegraphics[width=0.9\textwidth]{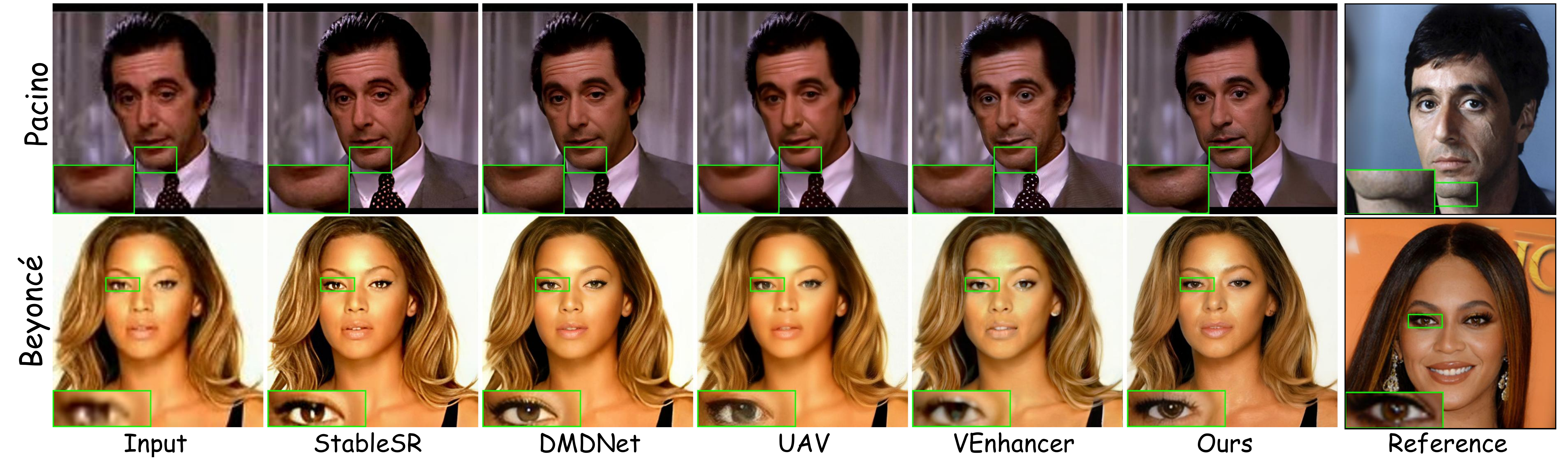}
\vspace{-0.5em}
\caption{Qualitative comparison on FOS-V. IP-FVR generates results with high identity preservation, capturing features like chin shape and iris color.}
\vspace{-1em}
\label{fig:quali_3}
\end{figure*}

%% file: Table/main_exp.tex
\begin{table}[t]
\centering
\caption{Quantitative comparisons of different face restoration methods on YouRef-heavy. The best and second performances are marked in \textcolor{red}{red} and \textcolor{blue}{blue}, respectively.}
\resizebox{\linewidth}{!}{
\begin{tabular}{lcccccccccc}
\toprule
Method & PSNR↑ & SSIM↑ & LPIPS↓ & CLIP-IQA↑ & MUSIQ↑ & LIQE↑ & IDS↑ & $\sigma_{IDS}$↓ & $E_{warp}$↓ \\
\midrule

DMDNet~\cite{li2022learning} 
& 28.35 & 0.850 & 0.206 & 0.593 & 68.82 & 3.980 & 0.732 & 3.056 & 6.849 \\

KEEP~\cite{feng2024kalman} 
& 27.65 & 0.842 & 0.215 & 0.607 & 70.51 & 3.740 & 0.681 & 2.795 & 6.228 \\

CodeFormer~\cite{zhou2022towards} 
& 28.67 & 0.873 & \textcolor{red}{0.193} & 0.454 & 63.81 & 3.107 & \textcolor{blue}{0.749} & \textcolor{blue}{2.598} & 6.315 \\

StableSR~\cite{wang2024exploiting} 
& \textcolor{blue}{29.03} & \textcolor{blue}{0.874} & \textcolor{blue}{0.203} & 0.442 & 53.59 & 2.102 & 0.726 & 2.624 & 9.344 \\

RVSR~\cite{chan2022investigating} 
& 26.70 & 0.801 & 0.319 & 0.539 & 65.40 & 2.811 & 0.723 & 2.719 & 8.032 \\

UAV~\cite{zhou2024upscale} 
& 27.82 & 0.834 & 0.275 & 0.589 & 64.80 & 3.14 & 0.658 & 3.478 & \textcolor{red}{5.723} \\

VEnhancer~\cite{he2024venhancer} 
& 20.35 & 0.726 & 0.296 & \textcolor{red}{0.680} & \textcolor{blue}{73.68} & \textcolor{blue}{4.019} & 0.624 & 3.495 & 6.766 \\

Ours 
& \textcolor{red}{29.51} & \textcolor{red}{0.918} & 0.216 & \textcolor{blue}{0.670} & \textcolor{red}{74.41} & \textcolor{red}{4.144} & \textcolor{red}{0.821} & \textcolor{red}{2.475} & \textcolor{blue}{5.802} \\
\bottomrule
\end{tabular}
}
\label{syn_data}
\end{table}

%% file: Table/rw_exp.tex
\begin{table}[t]

\centering
\caption{Quantitative comparisons of different face restoration methods based on FOS-V dataset. The best and second performances are marked in \textcolor{red}{red} and \textcolor{blue}{blue}, respectively.}

\resizebox{\linewidth}{!}
{
\begin{tabular}{lcccccc}
\toprule
 & CodeFormer~\cite{zhou2022towards} & StableSR~\cite{wang2024exploiting} & RVSR~\cite{chan2022investigating} & UAV~\cite{zhou2024upscale} & VEnhancer~\cite{he2024venhancer} & Ours \\ 
\midrule
CLIP-IQA↑ &\textcolor{blue}{0.452} &0.391 & 0.469 &0.424 &0.446 & \textcolor{red}{0.534} \\ 
MUSIQ↑ &54.83 &46.10 &\textcolor{blue}{61.15} &49.86 &48.20 &\textcolor{red}{62.50} \\ 
LIQE↑ &\textcolor{blue}{2.701} &1.682 &2.248 &1.697 &1.802 &\textcolor{red}{2.916} \\ \midrule
$E_{warp}$($\times 10^{-3}$)↓ &9.253 &10.49  &10.742 &\textcolor{blue}{6.493}  &7.924 &\textcolor{red}{6.317}  \\ 
\bottomrule
\end{tabular}
}
\vspace{-0.5cm}
\label{rw_data}
\end{table}

%% file: Figure/ab-1.tex
\begin{figure}[h]
\centering
\includegraphics[width=0.49\textwidth]{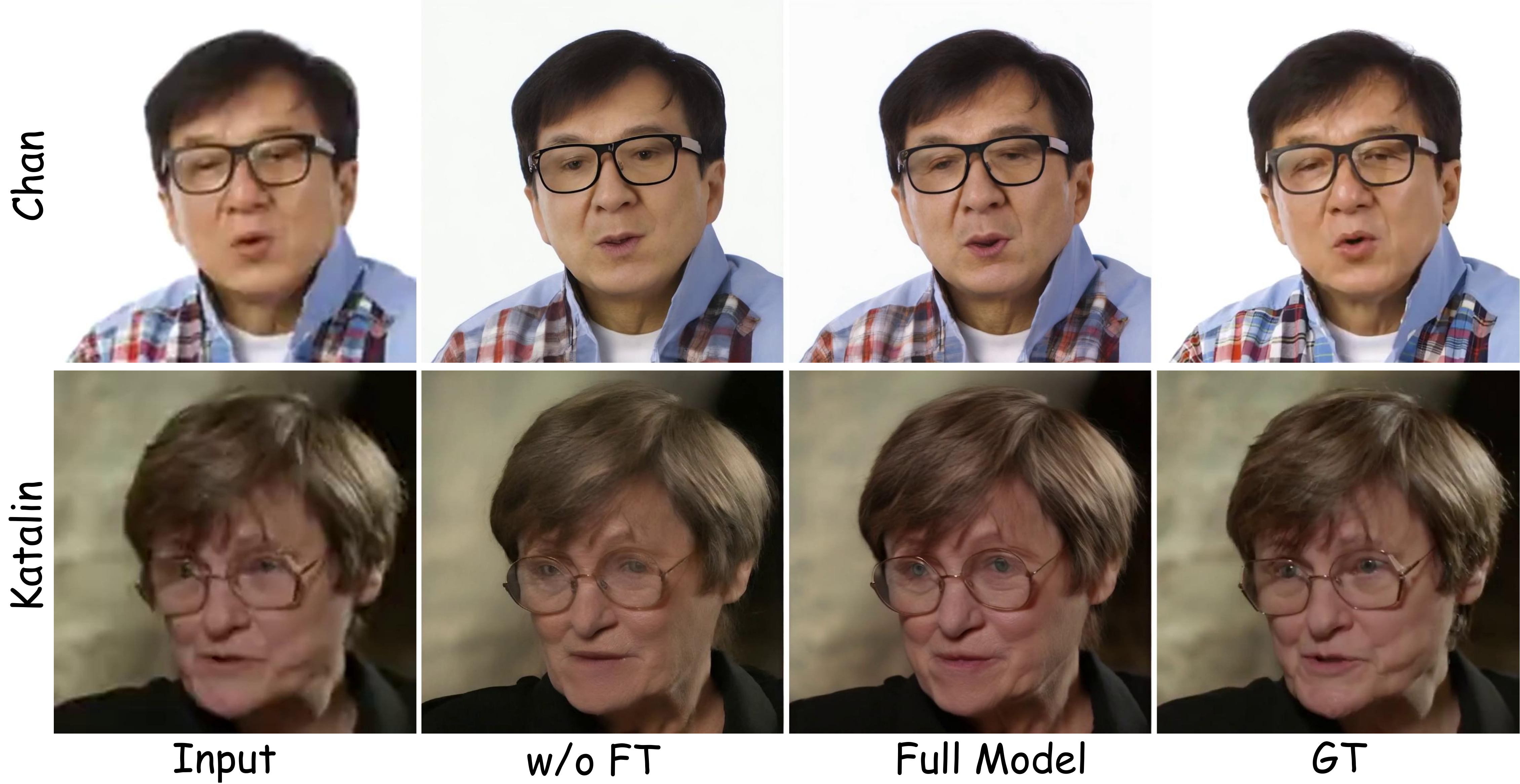}
\vspace{-1em}
\caption{A comparison of results before and after applying identity preserving feedback learning. Both approaches are capable of recovering basic identity features (\textit{e.g.}, nose shape and eye color), while personalized training further restores deeper characteristics, such as skin texture.}
\vspace{-1em}
\label{fig:ab-1}
\end{figure}

%% file: Figure/ab-2.tex
\begin{figure*}[h]
\centering
\includegraphics[width=0.9\textwidth]{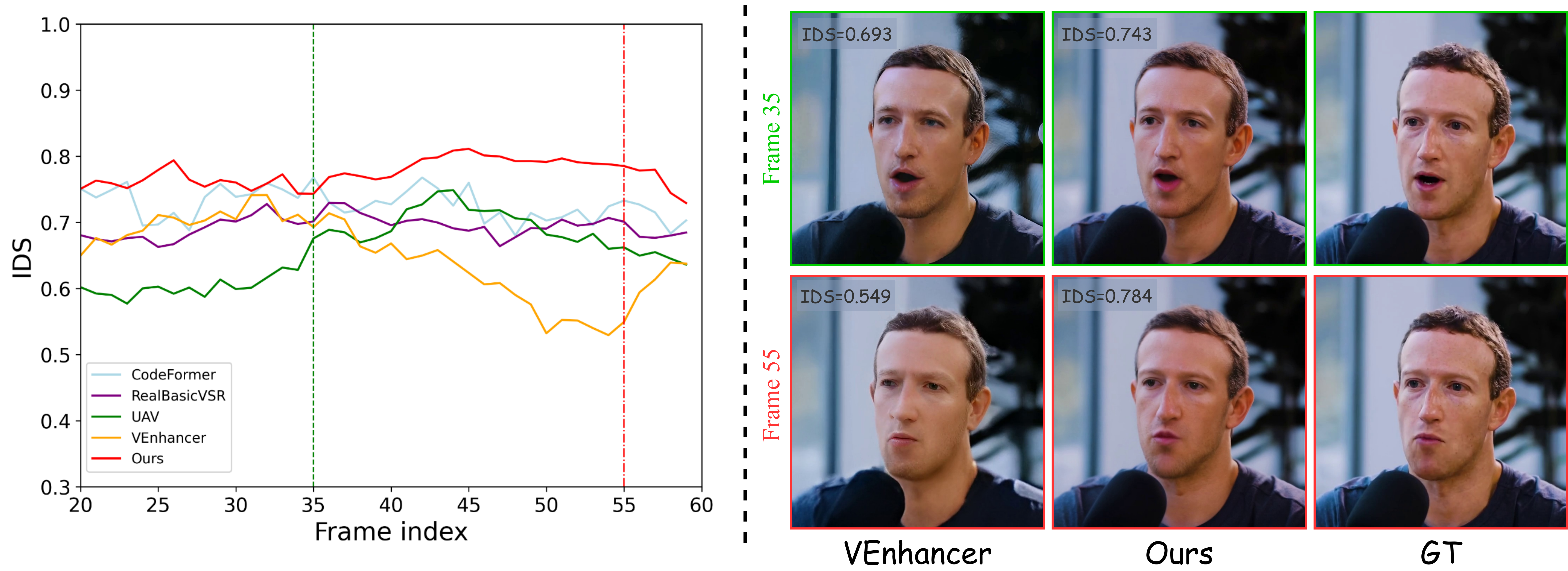}
\vspace{-1em}
\caption{Identity similarity across frames. Our method employs the Exponential Blending Strategy, effectively reducing the identity similarity fluctuations over time.}
\vspace{-1em}
\label{fig:cross_frame}
\end{figure*}

%% file: Table/tab-ab1.tex
% \begin{table}[h]
% \centering

% \caption{}
% \resizebox{\linewidth}{!}
% {
% \begin{tabular}{lccc}
% \toprule
% \textbf{Metrics} & \textbf{w/o.Face2Text Encoder } & \textbf{w/o.Visual Encoder} & \textbf{Full Model} \\
% \midrule
% PSNR↑ & 1000 &1000 &1000\\
% SSIM↑ & 1000 &1000 &1000\\
% LPIPS↓ & 1000 &1000 &1000\\
% \midrule
% IDS↑ & & & \\
% \bottomrule
% \end{tabular}
% }
% \label{tab:ab1}
% \end{table}

% \begin{table}[h!]
% \centering
% \resizebox{\linewidth}{!}{
% \begin{tabular}{cc|ccccccc}
% \toprule
% \multicolumn{2}{c|}{Prompts} & LPIPS $\downarrow$ & SSIM & PSNR & LIQE & CLIP-IQA & MUSIQ & IDS \\
% \cmidrule(lr){1-2} \cmidrule(lr){3-9}
% \textit{nq} & \textit{na} & & & & & & & \\
% \midrule
% \checkmark &  & 0.3264 & 0.6858 & 25.15 & 0.4782 & 0.2568 & 49.97 & \\
% \checkmark & \checkmark & 0.2690 & 0.6484 & 24.43 & 0.6441 & 0.7008 & 73.26 & \\
% \bottomrule
% \end{tabular}
% }
% \caption{Performance comparison of different prompt combinations across various quality metrics.}
% \label{tab:performance}
% \end{table}

\begin{table}[t]
    \centering
    \caption{Ablation Study of Decoupled Cross-Attention. The results indicate both modalities contribute to identity preservation.}
    \resizebox{\linewidth}{!}{
    \begin{tabular}{c c |ccccccc}
        \toprule
        \multicolumn{2}{c|}{Prompts} & 
        \multirow{2}{*}{PSNR$\uparrow$} & 
        \multirow{2}{*}{SSIM$\uparrow$} & 
        \multirow{2}{*}{LPIPS$\downarrow$} & 
        \multirow{2}{*}{CLIP-IQA$\uparrow$} & 
        \multirow{2}{*}{MUSIQ$\uparrow$} & 
        \multirow{2}{*}{LIQE$\uparrow$} & 
        \multirow{2}{*}{IDS$\uparrow$} \\
        
        \multicolumn{1}{c}{\textbf{Face2Text}} & \multicolumn{1}{c|}{\textbf{Visual}} & & & & & & & \\
        \midrule
        \checkmark & & 28.51 & 0.806 & 0.271 & 0.609 & 70.63 & 3.907 & 0.694 \\
        & \checkmark & 29.01 & 0.818 & 0.239 & 0.621 & 72.50 & 3.951 & 0.758 \\
        \checkmark & \checkmark & 29.51 & 0.918 & 0.216 & 0.670 & 74.41 & 4.144 & 0.821 \\
        \bottomrule
    \end{tabular}
    }
    \label{tab:ab1}
% \vspace{-0.5cm}
\end{table}

% \begin{table}[h]
%     \centering
%     \caption{Ablation Study of Decoupled Cross-Attention.}
%     \resizebox{\linewidth}{!}{
%     \begin{tabular}{c c | c c c c c c c}
%         \hline
%         \multicolumn{2}{c|}{Prompts} &PSNR$\uparrow$  & SSIM$\uparrow$ &LPIPS $\downarrow$    & CLIP-IQA & MUSIQ &LIQE &IDS \\
%         \hline
%         \textbf{Face2Text} & \textbf{Visual}  & 0.3264 & 0.6858 & 25.15 & 0.4782 & 0.2568 & 49.97 &\\
%         \checkmark  & & 0.2690 & 0.6484 & 24.43 & 0.6441 & 0.7008 & 73.26& \\
%         \checkmark & \checkmark  & 0.2930 & 0.6066 & 23.27 & 0.6656 & 0.7999 & 75.34& \\
%         \checkmark &  \checkmark & 0.2702 & 0.6511 & 24.49 & \textbf{\textcolor{red}{0.6437}} & 0.7029 & 73.11& \\
%         \hline
%     \end{tabular}
%     }

%     \label{tab:ab1}
% \end{table}

%% file: Table/tab-ab2.tex
\begin{table}[t]
\centering
\caption{Ablation Study of Exponential Blending, the results highlights the importance of both EB and NS in enhancing smooth transitions and identity stability.}
\resizebox{0.8\linewidth}{!}
{
\begin{tabular}{lccc}
\toprule
\textbf{Metrics} & \textbf{w/o NS\& EB} & \textbf{w/o EB} & \textbf{Full Model} \\
\midrule
$\sigma_{IDS}$($\times 10^{-2}$)↓ &3.016 &2.907 &2.475 \\
$E_{warp}$($\times 10^{-3}$)↓  &6.701 &6.351 &5.802 \\

\bottomrule
\end{tabular}
% \vspace{-1cm}
}
\label{tab:ab2}
\end{table}

%% file: Table/tab-ab3.tex
\begin{table}[t]
\centering
\label{tab:ab3}

\caption{Ablation Study of Negative Prompt. The results highlight the effectiveness of combining both negative text and visual prompts for improved identity preservation.}
\resizebox{0.84\linewidth}{!}{
\begin{tabular}{lcccc}
\toprule
\textbf{Metrics} & \textbf{w/o NT\&NV}& \textbf{w/o NT} & \textbf{w/o NV}  &\textbf{Full Model} \\
\midrule
PSNR↑ &27.12 &28.04 &\textbf{29.40} &29.23 \\
SSIM↑ &0.840 &0.865 &\textbf{0.914} &0.902 \\
LPIPS↓ &0.291 &0.264 &0.236 &\textbf{0.215} \\
\midrule
IDS↑ &0.760 &0.756 &0.769 &\textbf{0.781} \\
\bottomrule
\end{tabular}
}
\vspace{-0.5cm}
\label{tab:ab3}
\end{table}